\documentclass[11pt]{article}
\usepackage{amsmath} 
\usepackage{amssymb}
\usepackage[preprint]{acl}
\usepackage{times}
\usepackage{latexsym}
\usepackage[nointegrals]{wasysym} 
\usepackage{pifont}
\usepackage{graphicx}
\usepackage{caption} 
\usepackage[T1]{fontenc}
\usepackage[utf8]{inputenc}
\usepackage{microtype}
\usepackage{inconsolata}
\usepackage{booktabs}
\usepackage{multirow}
\usepackage{tabularx}
\usepackage{makecell}
\usepackage{array}
\usepackage{ragged2e}
\DeclareSymbolFont{extraup}{U}{zavm}{m}{n}
\DeclareMathSymbol{\varheart}{\mathord}{extraup}{86}
\newcommand{\MyHeart}{\raisebox{-0.2ex}{\scalebox{1.0}{\ensuremath{\varheart}}}}
\usepackage{enumitem}
\usepackage{siunitx}
\usepackage{stfloats}
\usepackage{hyperref}

\title{Fine-grained Verbal Attack Detection via a Hierarchical Divide-and-Conquer Framework}

\author{
Quan Zheng\textsuperscript{$\blacklozenge$$\clubsuit$} \quad
Yuanhe Tian\textsuperscript{\MyHeart$\clubsuit$\thanks{Corresponding author.}} \quad
\vspace{0.1cm}
Ming Wang\textsuperscript{$\blacklozenge$} \quad
Yan Song\textsuperscript{$\spadesuit$} \\
\textsuperscript{$\blacklozenge$}Beijing Normal University \quad \textsuperscript{$\clubsuit$}Zhongguancun Academy \\
\textsuperscript{\MyHeart}Zhongguancun Institute of Artificial Intelligence \\
\vspace{0.1cm}
\textsuperscript{$\spadesuit$}University of Science and Technology of China \\
\textsuperscript{$\blacklozenge$}\texttt{s-zq25@bjzgca.edu.cn} \quad
\textsuperscript{\MyHeart}\texttt{tianyuanhe@zgci.ac.cn} \\
\textsuperscript{$\blacklozenge$}\texttt{wangming@bnu.edu.cn} \quad
\textsuperscript{$\spadesuit$}\texttt{clksong@gmail.com}
}

\begin{document}
\maketitle

\begin{abstract}
In the digital era, effective identification and analysis of verbal attacks are essential for maintaining online civility and ensuring social security. However, existing research is limited by insufficient modeling of conversational structure and contextual dependency, particularly in Chinese social media where implicit attacks are prevalent. Current attack detection studies often emphasize general semantic understanding while overlooking user response relationships, hindering the identification of implicit and context-dependent attacks. To address these challenges, we present the novel “Hierarchical Attack Comment Detection” dataset and propose a divide-and-conquer, fine-grained framework for verbal attack recognition based on spatiotemporal information. The proposed dataset explicitly encodes hierarchical reply structures and chronological order, capturing complex interaction patterns in multi-turn discussions. Building on this dataset, the framework decomposes attack detection into hierarchical subtasks, where specialized lightweight models handle explicit detection, implicit intent inference, and target identification under constrained context. Extensive experiments on the proposed dataset and benchmark intention detection datasets show that smaller models using our framework significantly outperform larger monolithic models relying on parameter scaling, demonstrating the effectiveness of structured task decomposition.

\end{abstract}

\section{Introduction}
The advent of the digital era has revolutionized communication, offering unparalleled convenience while simultaneously introducing significant social security risks \cite{weinberg2015internet,strauss2024analyzing,odesanmi2025double}.
Notably, online verbal attack has become increasingly frequent and covert, posing a critical challenge to the governance of online environments \cite{kumar2018benchmarking,elsherief2021latent,hartvigsen2022toxigen}. 
The vast majority of online verbal attacks exhibit covert characteristics, making them difficult to identify based solely on their literal meaning \cite{wiegand2021implicitly,frenda2023killing,paakki2024detecting}.
Moreover, numerous internet slang terms provide a covert boost to verbal attacks, making them difficult to interpret correctly without the appropriate context and conversational setting \cite{mattiello2022language}.
However, the advent of large language models (LLMs) and their success in processing social media content \cite{xu2025consistency,zhang2024somelvlm,tian2025multimodal,liu2025balanced,guo2025sns} have opened up new avenues for identifying and analyzing cyberbullying \cite{kumar2024bias,shekhar2024comparative,cirillo2025exploring}.

Verbal attack recognition is a vital subfield of implicit intent understanding.
Current approaches fall into three categories: dictionary and rule-based, conventional machine learning, and deep learning methods \cite{fortuna2018survey,simon2022trends,farias2024proceedings}.
Dictionary and rule-based approaches use predefined lexicons but struggle with evolving expressions \cite{schmidt2017survey,davidson2017automated,pedersen2019duluth}.
Similarly, conventional machine learning classifiers often fail to capture complex semantic dependencies \cite{luts2014mean,djuric2015hate,badjatiya2017deep}.
In contrast, deep learning methods have become mainstream for modeling context and enhancing robustness \cite{lecun2015deep,young2018recent,devlin2019bert,sekkate2024deep}.
Despite this progress, limitations persist \cite{roy2023probing}.
First, current datasets lack linguistic diversity and often overlook reply tree structures \cite{vidgen2020directions,arshad2024understanding}.
Second, existing methods struggle with context, frequently introducing noise or missing fine-grained cues \cite{yoon2025pulmo,aleroud2016using}.
Finally, LLMs face efficiency bottlenecks compared to specialized micro-models offering better performance and lower costs \cite{wang2025false,su2025toolorchestra}.
Driven by these limitations, we require a framework that achieves precise contextual adoption via multiple specialized micro-models.

In this paper, we first construct a Chinese Hierarchical Attack Comment Detection dataset (HACD) sourced from social media platforms.
Building on this, we introduce spatiotemporal information encompassing hierarchical and chronological order, and propose a fine-grained detection and analysis framework for verbal attacks based on hierarchical divide-and-conquer.
The framework incorporates four smaller models functioning as explicit/implicit attack detectors and analyzers.
Considering task difficulty, the explicit attack models are smaller than the implicit ones.
This design lowers computational costs while maintaining accuracy by efficiently leveraging model capabilities.
We conduct comprehensive experiments on model training and frameworks.
Results show that smaller models using our framework achieve performance comparable to or exceeding large-scale models.
This empirically validates the effectiveness of our framework and spatiotemporal information.
The main contributions of this paper are as follows:
\begin{itemize}[leftmargin=*, nosep]
    \item \textbf{Dataset Construction:} We construct a Chinese comment dataset with a tree-structured hierarchy, labeled using human-supervised LLM techniques, providing a resource for verbal attack recognition research.
    \item \textbf{Spatiotemporal Information:} We propose and utilize hierarchical levels and chronological order to ensure precise contextual utilization, enabling the accurate analysis of attack behaviors across dynamic and complex social interactions.
    \item \textbf{Hierarchical Divide-and-Conquer framework:} We establish a fine-grained framework for verbal attack detection via a divide-and-conquer strategy, enhancing multi-level recognition capabilities while optimizing computational costs.
\end{itemize}

\section{HACD Construction}

\subsection{Data Collection and Processing}

\begin{figure}[t]
    \centering
    \includegraphics[width=0.45\textwidth,trim=0 0 0 0]{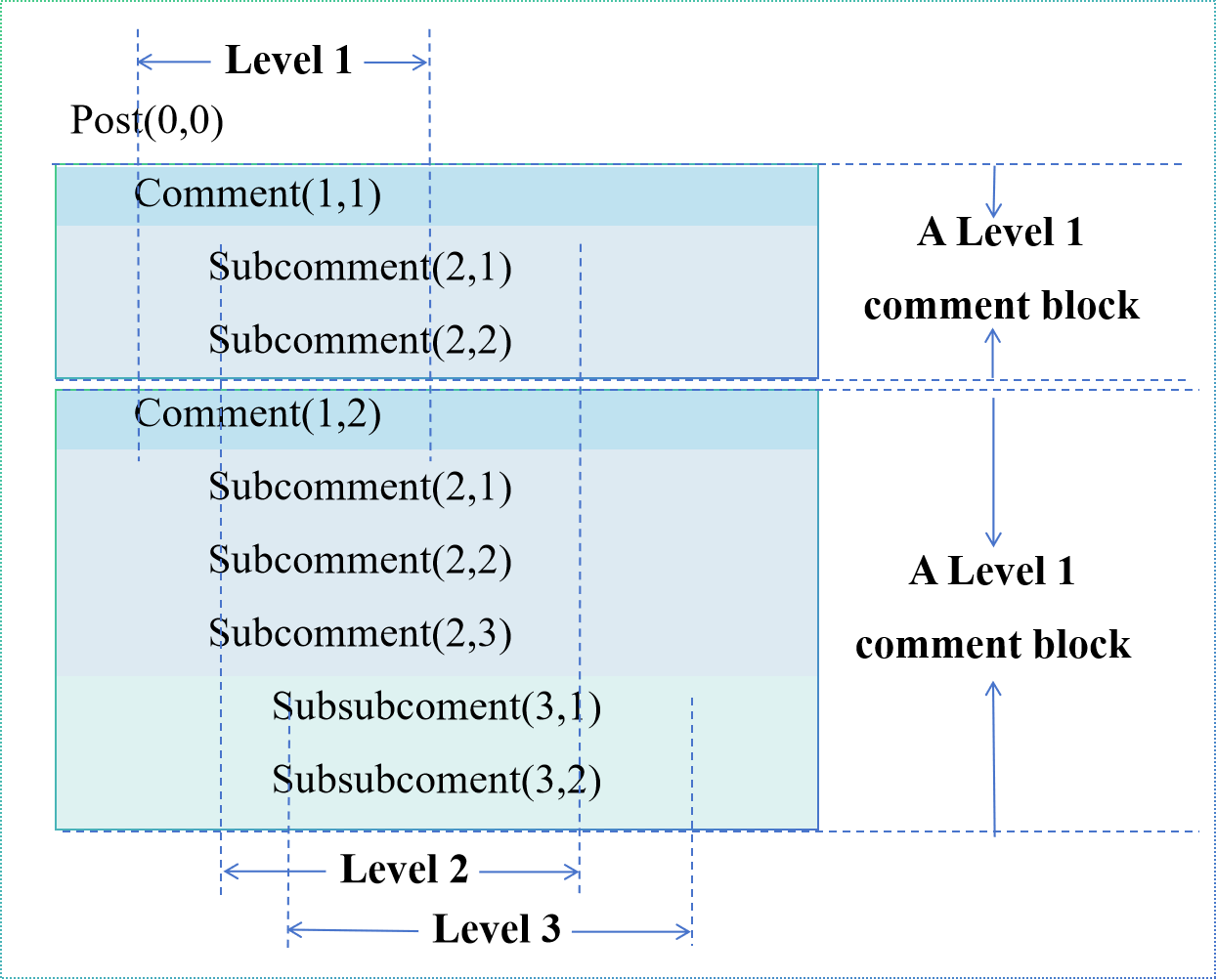}
    \caption{Data structure after preprocessing. Level 1 denotes a first-level comment, while Level 2 denotes a second-level comment. A level comment block represents a first-level comment block. (x,y) indicates that the comment belongs to level x and has a sequence number of y within that level. For example, Subcomment(2,1) in the figure refers to the first secondary comment under the primary comment Comment(1,1).}
    \label{fig:Comment Encoding Method}
    \vspace{-0.6cm}
\end{figure}

We primarily collect data from Weibo\footnote{\url{https://www.bilibili.com/}} and Bilibili\footnote{\url{https://weibo.com/}}.
Weibo is a very prominent Chinese social media platform offering information sharing, interactive commenting, and topic discussions.
Bilibili is an extremely well-known Chinese video community featuring video content, posts, and comment sections.
The comment sections on both platforms provide particularly natural tree-like reply structures.
To capture high user engagement and complex discussion trajectories, we focus on highly controversial videos addressing gender dynamics, AI ethics, and evaluations of historical figures.
To better simulate real-world user debates, we preserved the platforms' inherent tree-like reply hierarchy.
To ensure contextual integrity, we define the level 1 comment block as the smallest unit.
A level 1 comment block consists of a single First-level comment as the anchor, with its complete chain of sub-comments, as depicted in Figure \ref{fig:Comment Encoding Method}.
This structure encapsulates a specific discussion branch's lifecycle, isolating it from unrelated discourse in the broader section.
To accurately capture the discourse flow within these Thread Blocks, we incorporate the spatiotemporal information of comment texts—specifically, hierarchical level and chronological order that capture the structural depth and temporal sequence of the conversation.
The result of the data preprocessing is shown in Figure \ref{fig:Comment Encoding Method}.
Hierarchical level and chronological order are defined as follows:
\begin{table}[t]
    \centering
    \sisetup{detect-family, detect-weight, detect-shape}
    \begin{tabular}{l S[table-format=1.2] S[table-format=1.2]}
        \toprule
        \textbf{Label} & \textbf{Kappa} & \textbf{Consistency rate} \\
        \midrule
        Attack or not  & 0.91 & 0.90 \\
        Attack form    & 0.85 & 0.94 \\
        Attack target  & 0.92 & 0.96 \\
        Attack type    & 0.88 & 0.92 \\
        Attack intent & 0.82 & 0.90 \\
        Harmfulness    & 0.82 & 0.88 \\
        \bottomrule
    \end{tabular}
    \caption{Kappa test for dataset annotation. Kappa results comparing machine and human annotation for each feature. Kappa values and consistency rates both exceeding 0.8 indicate excellent consistency.}
    \label{tab:Kappa Test for Dataset Annotation}
    \vspace{-0.5cm}
\end{table}
\begin{table}[t]
    \footnotesize
    \setlength{\tabcolsep}{3pt}
    \centering
    \scalebox{0.97}{
    \begin{tabular}{@{}>{\raggedright\arraybackslash}m{1.3cm}>{\raggedright\arraybackslash}m{0.8cm}>{\raggedright\arraybackslash}m{2.5cm}>{\raggedleft\arraybackslash}m{0.8cm}>{\centering\arraybackslash}m{0.8cm}@{}}
        \toprule
        \textbf{Label} & \textbf{Type} & \textbf{Distribution} & \textbf{Count} & \textbf{Percent} \\
        \midrule
        
        \multirow{3}{=}{Attack or not} & \multirow{3}{=}{Class.} & Explicit attack & 9,275 & 35.1\% \\
        & & Implicit attack & 4,766 & 18.0\% \\
        & & No attack & 12,382 & 46.9\% \\
        \midrule
        
        \multirow{3}{=}{Attack form} & \multirow{3}{=}{Class.} & Targeted & 10,775 & 40.8\% \\
        & & Non-targeted & 6,574 & 24.9\% \\
        & & No attack & 9,074 & 34.3\% \\
        \midrule
        
        \multirow{3}{=}{Attack target} & \multirow{3}{=}{Class.} & Individuals & 10,258 & 38.8\% \\
        & & Group & 740 & 2.8\% \\
        & & No attack & 15,425 & 58.4\% \\
        \midrule
        
        \multirow{7}{=}{Attack type} & \multirow{7}{=}{Class.} & Discriminatory & 323 & 1.2\% \\
        & & Satirical & 2,372 & 9.0\% \\
        & & Abusive & 385 & 1.5\% \\
        & & Threat & 24 & 0.1\% \\
        & & Demeaning & 5,481 & 20.7\% \\
        & & Others & 5,433 & 20.6\% \\
        & & No attack & 12,405 & 47.0\% \\
        \midrule
        
        \multirow{10}{=}{Attack intent} & \multirow{10}{=}{Class.} & Racism & 99 & 0.4\% \\
        & & Gender dichotomy & 142 & 0.5\% \\
        & & Hate speech & 317 & 1.2\% \\
        & & Personal attacks & 3,288 & 12.4\% \\
        & & Verbal mockery & 55 & 0.2\% \\
        & & Personal insults & 101 & 0.4\% \\
        & & Stereotypes & 493 & 1.9\% \\
        & & Security threat & 12 & 0.1\% \\
        & & Others & 4,583 & 17.4\% \\
        & & No attack & 17,331 & 65.6\% \\
        \midrule
        
        \multirow{3}{=}{Hazard Level} & \multirow{3}{=}{Num.} & Average & 9.89 & - \\
        & & Std. dev. & 13.82 & - \\
        & & Max & 80.00 & - \\
        \midrule
        
        \multirow{3}{=}{Confidence Level} & \multirow{3}{=}{Num.} & Average & 86.58 & - \\
        & & Std. dev. & 6.51 & - \\
        & & Max & 100.00 & - \\
        \bottomrule
    \end{tabular}}
    \caption{Detailed statistics of the dataset label distribution across multiple dimensions. The dataset is annotated with five classification tasks ranging from coarse-grained detection (e.g., Attack or not) to fine-grained analysis (e.g., Attack type and intent). The bottom rows present numerical statistics (Mean, Std. Dev., Max) for the Hazard and Confidence levels.}
    \label{tab:label_distribution}
    \vspace{-0.6cm}
\end{table}
\begin{itemize}[leftmargin=*, nosep]
\item \textbf{Hierarchical level:} This marks the vertical depth of a reply within the conversation tree. The root comment is designated as Level-1. A comment that responds directly to the root is encoded as Level-2, while a reply to a Level-2 node is encoded as Level-3, and so forth.
\item \textbf{Chronological order:} Capturing the horizontal sequence, comments at the same level are indexed by timestamp for temporal consistency.
\end{itemize}

\subsection{Data Annotation}
After collecting and preprocessing the data, we proceed with data annotation. 
Specifically, we first design the labeling system, then develop an annotation workflow leveraging large language models, and finally control annotation quality through manual verification. 
Details are outlined below.

\paragraph{Hierarchical taxonomy formulation}
To comprehensively analyze attack behaviors within tree-structured discussions, we design a multi-dimensional hierarchical taxonomy by drawing upon linguistic perspectives \cite{zampieri2019semeval,bkaczkowska2024implicit}, target-based classifications \cite{bourgonje2017automatic}, and attack forms \cite{kogilavani2023characterization}.
The annotation schema classifies comments across five granularity levels:
\begin{itemize}[leftmargin=*, nosep]
\item \textbf{Attack or not:} Identifies attack, distinguishing between \textit{explicit attack}, \textit{implicit attack}, and \textit{no attack} \cite{bkaczkowska2024implicit}.
\item \textbf{Attack form:} Identifies whether an attack is targeted at a specific entity, including \textit{targeted}, \textit{non-targeted}, and \textit{no attack} \cite{zampieri2019semeval}.
\item \textbf{Attack target:} Specifies the targets of the attack, categorizing them as \textit{individuals}, \textit{group} or \textit{no attack} \cite{zampieri2019semeval}.
\item \textbf{Attack type:} Classifies the nature of offensive language into fine-grained categories such as \textit{discriminatory}, \textit{satirical}, \textit{abusive}, \textit{threat}, \textit{demeaning}, \textit{other}, and \textit{no attack} \cite{kogilavani2023characterization}.
\item \textbf{Attack intent:} Analyzes the underlying motivation, covering specific intents like \textit{racism}, \textit{sexism}, \textit{hate speech}, \textit{personal attacks}, \textit{verbal mockery}, \textit{personal insults}, \textit{stereotypes}, \textit{safety threats}, \textit{other}, \textit{no attack} \cite{bourgonje2017automatic}.
\end{itemize}
Additionally, we incorporate numerical indicators for Hazard level and Confidence level to quantify severity and annotation certainty, respectively.

\paragraph{LLM-assisted annotation workflow} 
Motivated by recent evidence that state-of-the-art LLMs can act as reliable and cost-effective annotators for supervised data construction \cite{ding2023gpt,gilardi2023chatgpt,zhu2025trove,wang2025define}, we employ the GPT-4 API\footnote{\url{https://platform.openai.com/docs/guides/text}}
 as the primary annotator, leveraging its strong capability in contextual and pragmatic understanding \cite{achiam2023gpt}.
To adapt the model to domain-specific discourse, we perform prompt engineering with carefully curated few-shot examples drawn from Bilibili and Weibo comment threads.
During annotation, GPT-4 is instructed to evaluate each target comment strictly within the scope of its corresponding first-level comment block, ensuring that judgments—particularly for implicit attacks—are grounded in conversationally available context rather than isolated utterances.

\paragraph{Quality assurance and human verification} 
To rigorously validate the reliability of our LLM-generated annotations, we conduct a comprehensive consistency analysis. 
We randomly sample 200 representative data points for independent annotation by five human annotators. 
These annotators are graduate students specializing in linguistics, journalism and communication, and computer science.\footnote{All annotators are trained with detailed specifications.}
We calculate Cohen’s Kappa \cite{cohen1960coefficient} coefficient between the GPT-4 predictions and human consensus to quantitatively measure inter-rater reliability.
The inter-rater reliability results are presented in Table \ref{tab:Kappa Test for Dataset Annotation}.
These results show high Kappa values across all major taxonomic levels (e.g., \textit{Attack or not}, \textit{Attack form}), with an average of 0.87. This demonstrates substantial agreement and confirms the model's proficiency in aligning with human judgment on complex discursive nuances.
To obtain a portion of high-quality data, we manually annotate 2,607 data points and utilize 200 of them in the kappa test. This dataset will be employed in subsequent analyses.
\begin{figure*}[t]
    \centering
    \includegraphics[width=1.0\textwidth,trim=0 10 0 10]{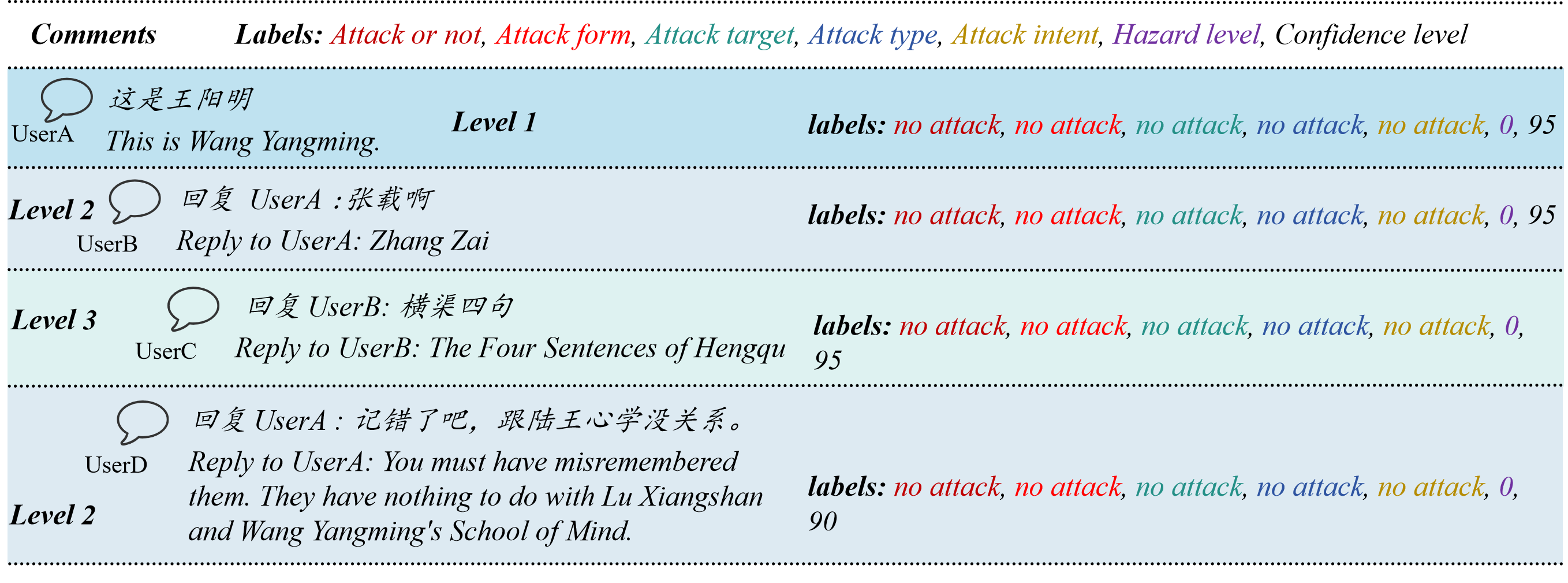}
    \caption{An illustrative example of an annotated conversation thread. The left side displays the hierarchical discourse structure (from Level 1 to Level 3), while the right side presents the corresponding fine-grained labels across seven dimensions, including attack form, target, type, intent, and hazard level.}
    \label{fig:data case32}
    \vspace{-0.5cm}
\end{figure*}

\subsection{The Properties of HACD}

After annotation, we obtain 25,993 data points, where 23,386 and 2,607 of them are annotated by LLM and annotators, respectively.\footnote{Our annotation indicates that there is no personal information in the data.}
The distribution of labels is shown in Table \ref{tab:label_distribution}. In the ``Attack or not'' category, ``no attack'' accounts for the largest proportion, followed by explicit attacks. Regarding ``Attack form'', direct targeted attacks represent the largest portion, exceeding non-attacking comments. Furthermore, ``no attack'' remains the dominant label across ``Attack target'', ``Attack type'', and ``Attack intent''. The distribution is balanced and reflects real-world scenarios.
Thanks to inherent reply relationships and hierarchical depth, our dataset facilitates not only identifying and analyzing verbal attack but also tasks such as constructing dialogue chains and generating user profiles, demonstrating application prospects.
We use the 2,607 manually annotated points as the test set, with the remaining 23,386 allocated to training and validation, ensuring an 80:10:10 ratio.
A random sample from the dataset is shown in Figure \ref{fig:data case32}.

\begin{figure*}[t]
    \centering
    \includegraphics[width=1.0\textwidth]{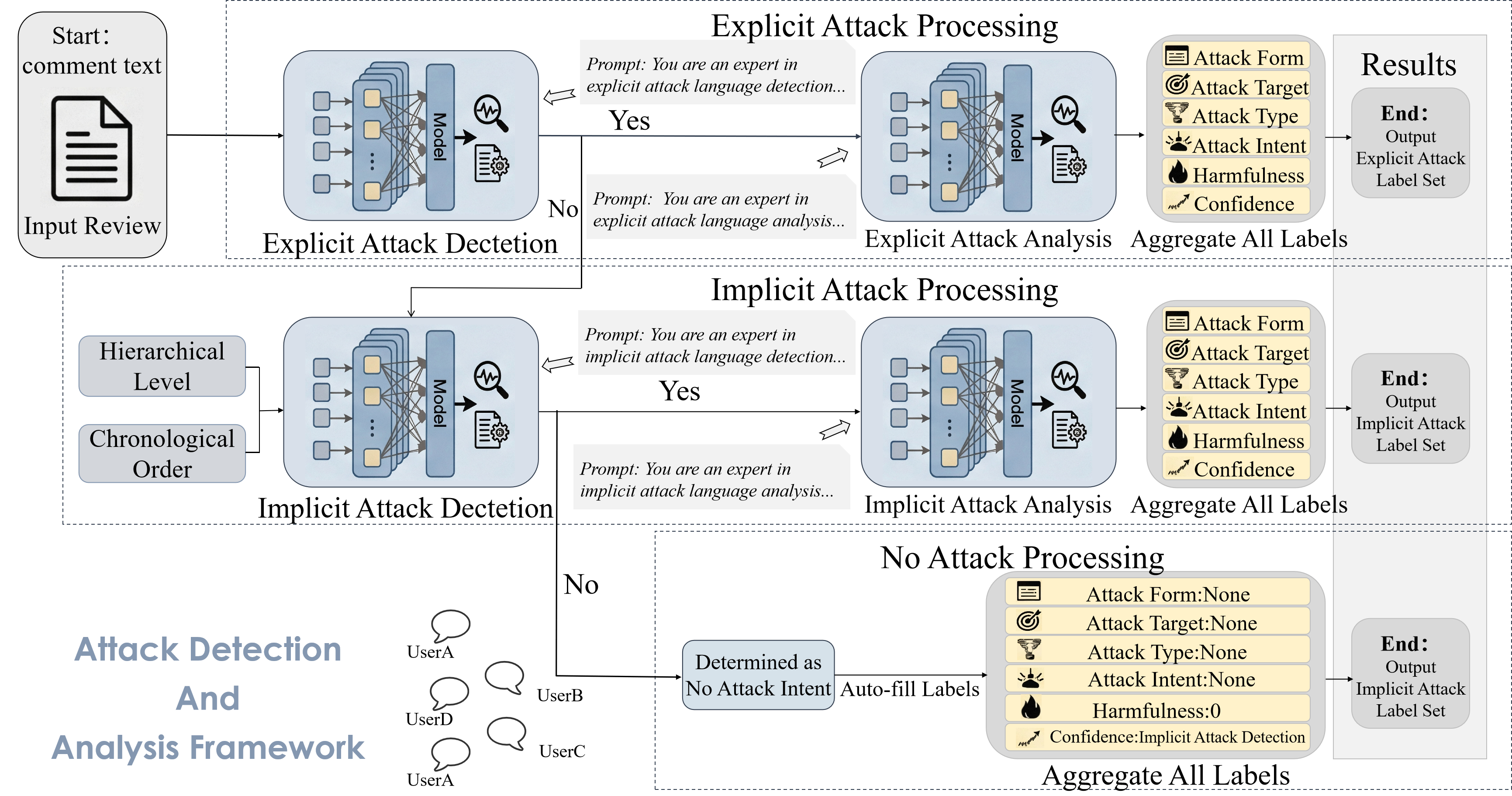}
    \caption{Overview of the proposed attack detection and analysis framework. The processing pipeline first evaluates comment text for explicit attack. If such markers are absent, it incorporates structural context to detect implicit attack. All detected attacks are then routed to specialized analyzers to generate fine-grained labels (e.g., target, intent, harmfulness), while any identified non-attacks are automatically labeled with null values.}
    \label{fig:Attack Detecting and analysis Framework}
    \vspace{-0.3cm}
\end{figure*}

\section{The Approach}
To address the complexity of verbal attacks, we propose a hierarchical divide-and-conquer framework for granular detection and analysis, comprising four specialized lightweight models.
This framework differentiates between explicit and implicit attacks through a step-by-step workflow, as illustrated in Figure \ref{fig:Attack Detecting and analysis Framework}. 
The framework is primarily divided into four modules: explicit attack detector, explicit attack analyzer, implicit attack detector, and implicit attack analyzer. Each module is handled by a specially trained model. We integrate them through a pipeline to work collaboratively. 
The details of the aforementioned process are illustrated as follows.

\subsection{The Context Selection}
Following the preceding data preprocessing, we propose a controllable context selection mechanism to formalize our model's input.
For any target comment \textit{C}, its valid context set \textit{context} is defined as the union of its ancestors and its preceding siblings:
\begin{equation}
\setlength{\abovedisplayskip}{5pt}
\setlength{\belowdisplayskip}{5pt}
\scalebox{0.78}{$context = \{ \text{Ancestors}(C) \} \cup \{ \text{Preceding Siblings}(C) \}$}
\label{eq:1}
\end{equation}
where { Ancestors(\textit{C}) } includes all higher-level parent nodes on the path to the root, and { Preceding Siblings(\textit{C}) } includes strictly earlier comments at the same level.
This approach explicitly captures the tree-structured relationships while filtering out noise from parallel but irrelevant branches or future replies, ensuring the model reasons solely based on logically available information.

\subsection{Explicit Attack Detection and Analysis}
\paragraph{Explicit attack detection} 
Given the explicit attack intent of display attacks, which is able to be determined from the comment itself, introducing excessive context may introduce noise and impair the model's judgment.
Therefore, we provide only the comment itself (denoted as $comment$) to the explicit attack detector $f_{exd}$ for analysis and obtain the result $check_1$.
This process is formulated as:
\begin{equation}
\setlength{\abovedisplayskip}{5pt}
\setlength{\belowdisplayskip}{5pt}
\scalebox{0.86}{$check_1 = f_{exd}(comment)$}
\label{eq:2}
\end{equation}
\paragraph{Explicit attack analysis} 
For text that has been identified as explicit attack, different form the detection process, this module needs to conduct a comprehensive analysis of the target text $comment$, including the attack target, attack type, and other aspects.
We provide the output $check_1$ from the explicit attack detector $f_{exd}$ and the target text $comment$ to the explicit attack analyzer $f_{exa}$, and then obtain the analysis result $detail_1$.
This process is formulated as:
\begin{equation}
\setlength{\abovedisplayskip}{5pt}
\setlength{\belowdisplayskip}{5pt}
\scalebox{0.86}{$detail_1 = f_{exa}(check_1, comment)$}
\label{eq:4}
\end{equation}
Additionally, considering the varying difficulty of tasks and the barrel theory, we aim to ensure the overall performance of the framework while minimizing computational costs. We restrict the model sizes handled by explicit attack detectors and analyzers to be no larger than those managed by implicit attack detectors and analyzers. This approach prevents both wasted computational capacity and unnecessary increases in computational expenses.

\subsection{Implicit Attack Detection and Analysis}

\paragraph{Implicit attack detection} 
The attributes of implicit attacks, such as attack intent, targets, and forms, are often concealed beneath literal meanings.
Relying solely on the comment itself makes it difficult to identify and understand its attack intent.
Furthermore, to accurately interpret user intent, it is necessary to incorporate appropriate context from the user's subjective perspective, including contextual information and dialogue details.
We combine the previously mentioned hierarchical levels and chronological order by incorporating all parent comments and sibling comments posted earlier than the target comment, as shown in Eq.\eqref{eq:1}.
Additionally, to fully leverage the analysis results from the explicit attack detector, the implicit attack detector $f_{imd}$ takes the output $check_1$ from the explicit attack detector $f_{exd}$, the target text $comment$, and the context selected based on spatiotemporal information $context$ as inputs, and obtains the detection result $check_2$. This process is formulated as:
\begin{equation}
\setlength{\abovedisplayskip}{5pt}
\setlength{\belowdisplayskip}{5pt}
\scalebox{0.86}{$check_2 = f_{imd}(check_1, context, comment)$}
\label{eq:3}
\end{equation}
\paragraph{Implicit attack analysis} 
Similar to explicit attack analyzers, this module requires analyzing the target text across specific dimensions. 
The difference lies in the fact that this module processes text already identified as implicit attacks, presenting a higher level of difficulty. 
Therefore, we provide the model with appropriate contextual information and the analysis results from the implicit attack detector $f_{imd}$.
The implicit analyzer $f_{ima}$ inputs the output $check_2$ from the implicit attack detector $f_{imd}$, the target text $comment$, and the context selected based on spatiotemporal information $context$ as inputs and then obtains the analysis result $detail_2$. This process is formulated as:
\begin{equation}
\setlength{\abovedisplayskip}{5pt}
\setlength{\belowdisplayskip}{5pt}
\scalebox{0.86}{$detail_2 = f_{ima}(check_2, context, comment)$}
\label{eq:5}
\end{equation}

\begin{table}[t]
\centering
\setlength{\tabcolsep}{16pt}
\begin{tabular}{@{}lcc@{}}
\toprule
\textbf{Category} & \textbf{Count} & \textbf{Percentage (\%)} \\
\midrule
is\_abusive       & 6,047          & 20.29 \\
not\_abusive      & 23,762         & 79.71 \\
is\_explicit      & 5,597          & 18.78 \\
is\_implicit      & 450            & 1.51 \\
\bottomrule
\end{tabular}
\caption{Distribution of the InToxiCat (Catalan).}
\label{tab:InToxiCat (Catalan) distribution}
\vspace{-0.2cm}
\end{table}

\begin{table}[t]
\centering
\begin{tabular}{@{}lcc@{}}
\toprule
\textbf{Category} & \textbf{Count} & \textbf{Percentage (\%)} \\
\midrule
Hate Speech       & 1,430          & 5.77 \\
Offensive Language& 19,190         & 77.43 \\
None              & 4,163          & 16.80 \\
\bottomrule
\end{tabular}
\caption{Distribution of Hate Speech and Offensive Language (English).}
\label{tab:Offensive Language (English) distribution}
\vspace{-0.4cm}
\end{table}

\section{Experiment Settings}
\begin{table}[t]
    \centering
    \small
    \setlength{\tabcolsep}{4pt}
    \begin{tabular}{@{}lccccccccc@{}}
        \toprule
        Strategy & 0.5B & 1.5B & 4B & 6B & 8B & 9B & 11B & 14B & 24B \\
        \midrule
        NTNF & \checkmark &  &  &  & \checkmark\checkmark &  &  & \checkmark &  \\
        WTNF & \checkmark & \checkmark & \checkmark &  & \checkmark &  &  & \checkmark &  \\
        NTWF &  &  & \checkmark &  &  &  &  &  &  \\
        WTWF &  &  & \checkmark & \checkmark &  & \checkmark & \checkmark &  & \checkmark \\
        \bottomrule
    \end{tabular}
    \caption{Summary of model scales deployed for each experimental strategy. The table details the specific parameter sizes ranging from 0.5B to 24B used to validate different frameworks. The dual marks at the 8B scale represent concurrent experiments using two distinct backbones: Qwen and Llama.}
    \label{tab:experment setting}
    \vspace{-0.4cm}
\end{table}

\subsection{Datasets}
To evaluate the performance and cross-lingual generalization of our approach, we employ three datasets:
\begin{itemize}[leftmargin=*, nosep]
    \item \textbf{HACD (Chinese):} Our constructed dataset sourced from Bilibili and Weibo platforms, featuring a tree-structured reply hierarchy.
    \item \textbf{InToxiCat (Catalan):} A benchmark dataset for abusive language detection \cite{gonzalez2024building}.
    \item \textbf{Hate Speech and Offensive Language (English):} Widely used for detecting hate speech and offensive content \cite{davidson2017automated}.
\end{itemize}
The distributions of the two datasets, InToxiCat (Catalan) and Hate Speech and Offensive Language (English), are shown in Table \ref{tab:InToxiCat (Catalan) distribution} and Table \ref{tab:Offensive Language (English) distribution}, respectively, for clearer reference.
For all three datasets, we apply a random split of 80:10:10 for the training, validation, and test sets, respectively.

In addition, it is worth mentioning that before conducting specialized training for specific tasks, we pre-partition the dataset into four independent sub-datasets tailored for the four module models to learn. Based on this, we could conduct specialized training for each model, which is why our models can each leverage their strengths.

\begin{table*}[t!]
    \centering
    \small
    \setlength{\tabcolsep}{5pt}
    \sisetup{
        detect-mode,
        detect-family,
        table-format=1.2,
        table-number-alignment=center,
    }
    \scalebox{0.95}{
    \begin{tabular}{@{}
        >{\raggedright\arraybackslash}m{2.5cm}
        *{15}{S}
        @{}}
        \toprule
        \textbf{Metric} & 
        \multicolumn{4}{c}{\textbf{NTNF}} & 
        \multicolumn{1}{c}{\textbf{NTWF}} & 
        \multicolumn{5}{c}{\textbf{WTNF}} & 
        \multicolumn{5}{c}{\textbf{WTWF}} \\
        \cmidrule(lr){2-5} \cmidrule(lr){6-6} \cmidrule(lr){7-11} \cmidrule(lr){12-16}
        & 
        {\makecell{0.5B}} & 
        {\makecell{8B\\(Q)}} & 
        {\makecell{8B\\(L)}} & 
        {\makecell{14B}} & 
        {\makecell{4B}} & 
        {\makecell{0.5B}} & 
        {\makecell{1.5B}} & 
        {\makecell{4B}} & 
        {\makecell{8B}} & 
        {\makecell{14B}} & 
        {\makecell{4B}} & 
        {\makecell{6B}} & 
        {\makecell{9B}} & 
        {\makecell{11B}} &
        {\makecell{24B}} \\
        \midrule
        
        Attack\_or\_not\_Acc & 0.13 & 0.68 & 0.34 & 0.51 & 0.48 & 0.45 & 0.71 & 0.72 & 0.69 & 0.71 & 0.72 & 0.71 & 0.74 & 0.76 & \textbf{0.79} \\
        Attack\_or\_not\_F1 & 0.06 & 0.61 & 0.15 & 0.36 & 0.13 & 0.10 & 0.63 & 0.66 & 0.62 & 0.65 & 0.65 & 0.63 & 0.68 & \textbf{0.71} & 0.67 \\
        Attack\_form\_Acc & 0.10 & 0.61 & 0.48 & 0.47 & 0.34 & 0.35 & 0.70 & 0.74 & 0.70 & 0.73 & \textbf{0.77} & 0.70 & 0.74 & 0.75 & \textbf{0.77} \\
        Attack\_form\_F1 & 0.03 & 0.59 & 0.13 & 0.33 & 0.15 & 0.23 & 0.67 & 0.61 & 0.59 & 0.59 & 0.64 & 0.67 & 0.71 & 0.72 & \textbf{0.76} \\
        Attack\_target\_Acc & 0.56 & 0.65 & 0.64 & 0.72 & 0.64 & 0.59 & 0.81 & 0.76 & 0.77 & 0.78 & 0.84 & 0.81 & 0.84 & 0.85 & \textbf{0.87} \\
        Attack\_target\_F1 & 0.06 & 0.49 & 0.07 & 0.52 & 0.07 & 0.25 & 0.39 & 0.63 & 0.55 & 0.60 & 0.67 & 0.39 & 0.69 & \textbf{0.72} & 0.69 \\
        Attack\_type\_Acc & 0.19 & 0.50 & 0.48 & 0.48 & 0.49 & 0.46 & 0.64 & 0.60 & 0.58 & 0.60 & 0.69 & 0.64 & 0.67 & 0.70 & \textbf{0.74} \\
        Attack\_type\_F1 & 0.00 & 0.20 & 0.06 & 0.08 & 0.06 & 0.05 & 0.47 & 0.36 & 0.39 & 0.39 & 0.54 & 0.47 & 0.51 & 0.55 & \textbf{0.60} \\
        Attack\_intent\_Acc & 0.50 & 0.51 & 0.64 & 0.64 & 0.64 & 0.64 & 0.73 & 0.62 & 0.73 & 0.74 & 0.77 & 0.73 & 0.76 & 0.78 & \textbf{0.81} \\
        Attack\_intent\_F1 & 0.00 & 0.12 & 0.00 & 0.19 & 0.00 & 0.03 & 0.50 & 0.42 & 0.37 & 0.37 & 0.54 & 0.50 & 0.52 & 0.56 & \textbf{0.60} \\
        Harm\_Pearson & 0.65 & 0.54 & 0.22 & 0.22 & 0.09 & 0.64 & 0.66 & 0.57 & 0.60 & 0.66 & 0.64 & 0.73 & 0.74 & 0.80 & \textbf{0.86} \\
        Conf\_Pearson & -0.04 & 0.31 & 0.07 & 0.34 & 0.07 & 0.73 & 0.53 & 0.59 & 0.62 & 0.65 & 0.83 & 0.86 & 0.85 & \textbf{0.89} & 0.86 \\
        All\_in\_One\_Acc & 0.00 & 0.28 & 0.13 & 0.24 & 0.13 & 0.31 & 0.39 & 0.42 & 0.44 & 0.45 & 0.40 & 0.43 & 0.45 & 0.48 & \textbf{0.52} \\
        \bottomrule
    \end{tabular}}
    \vspace{-0.1cm}
    \caption{Detailed experimental results across different model scales (from 0.5B to 24B) and strategies. The 8B(Q) and 8B(L) denote Qwen-8B and Llama-8B, respectively. Metrics include Accuracy (Acc), F1-score, and Pearson correlation. Bold values indicate the highest performance achieved in each row.}
    \label{tab:Experimental Results}
    \vspace{-0.5cm}
\end{table*}

\subsection{Experiment Design and Implementation}
We design four comparative experimental settings to thoroughly validate the effectiveness of our specialized detection and analysis framework.
The configurations are listed in Table \ref{tab:experment setting}.

\paragraph{No train and no framework}
No train and no framework (NTNF) is the zero-shot baseline. 
We conduct inference using LLM without task-specific fine-tuning. The models include Qwen2.5-0.5B \cite{team2024qwen2}, Qwen3-8B \cite{yang2025qwen3}, Qwen3-14B, and Llama3-8B \cite{dubey2024llama}.

\paragraph{With train and no framework} 
With train and no framework (WTNF) uses the data directly for model training without segmentation and employs a single model per run without our framework. 
Models include Qwen2.5-0.5B, Qwen2.5-1.5B, Qwen3-4B, Qwen3-8B, and Qwen3-14B.

\paragraph{No train and with framework}
No train and with framework(NTWF) evaluates our modular pipeline in a zero-shot manner.
We assign specific lightweight models to different components: Qwen2.5-0.5B is deployed as the explicit attack detector and analyzer, while Qwen2.5-1.5B serves for implicit attacks.
The aggregate parameter count for this configuration is approximately 4B.

\paragraph{With train and with framework}
With train and with framework(WTWF) represents our full approach.
We reorganize the dataset to create task-specific sub-datasets for training the four modules (explicit/implicit detection and analysis).
We conduct extensive experiments using various combinations of the Qwen series models, resulting in framework configurations with parameter counts of 4B, 6B, 9B, 11B, and 24B.
The 4B model consists of two 0.5B models and two 1.5B models.
The 6B model comprises four 1.5B models.
The 9B model combines two 0.5B models and two 4B models.
The 11B model integrates two 1.5B models and two 4B models.
The 24B model incorporates two 4B models and two 8B models.
\section{Results and Analysis}
\subsection{Overall Performance Comparison}
Table \ref{tab:Experimental Results} presents the vertical comparison results.
First, the results reveal a consistent hierarchy: WTWF > WTNF > NTWF > NTNF. This ranking underscores the dual benefits of supervised fine-tuning and our specialized architecture, indicating they contribute cumulatively to effectiveness. Second, under identical training conditions, incorporating the proposed framework yields significant gains (i.e., WTWF > WTNF, and NTWF > NTNF), further validating that our modular divide-and-conquer strategy consistently enhances attack recognition capabilities across diverse settings.

Regarding refined tasks, we observe:
first, for identification and targeting, WTWF-24B shows dominance with the highest accuracy in detection, form, and target tasks. Meanwhile, WTWF-11B achieves a slightly higher detection F1-score. This suggests that while the largest model excels in precision, the 11B variant maintains competitive sensitivity and remains robust in borderline cases.
Second, in multi-label type and intent analysis, WTWF-24B leads with the highest Subset Accuracy and F1 scores. This indicates larger parameter capacities better capture complex semantic features and long-tail distributions, particularly in nuanced or ambiguous attack scenarios.
Third, concerning consistency, WTWF-24B shows the strongest correlation in hazard assessment, whereas WTWF-11B outperforms in confidence assessment.
Overall, WTWF-24B attains the highest All\_in\_One\_Accuracy, followed closely by WTWF-11B, demonstrating stable improvements across almost all evaluation dimensions.
These results align with scaling laws, yet the gains are not solely attributable to parameter capacity. The improvements also stem from explicit task separation and context control, suggesting that architectural decomposition provides complementary benefits beyond model scaling and offers a more interpretable pathway for performance growth.

\subsection{Framework Efficiency Analysis}

\begin{figure}[t]
    \scalebox{0.95}{
    \centering
    \includegraphics[width=0.5\textwidth]{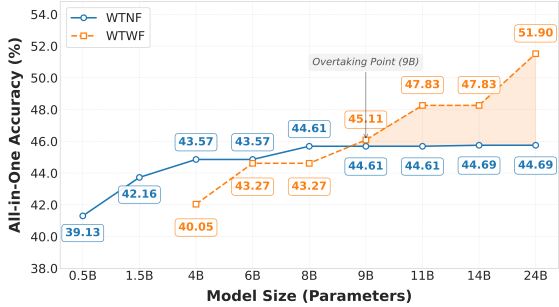}
    }
    \caption{Performance comparison of WTNF vs. WTWF strategies across model sizes. The shaded orange area highlights WTWF's accuracy gain over WTNF after overtaking at 9B parameters.}
    \label{fig:WTWF vs WTNF Performance Improvement}
    \vspace{-0.3cm}
\end{figure}

To investigate model scaling and framework efficacy, we plot the performance in Figure \ref{fig:WTWF vs WTNF Performance Improvement}.
There are several observations regarding scalability.
First, WTWF's performance growth significantly outpaces that of WTNF as parameter size increases (widening the gap by approximately 21\%).
This demonstrates our framework's superior scalability, effectively converting increased parameters into performance gains compared to the baseline.
Second, saturation behaviors are observed: while the performance improvement of the WTNF group begins to stagnate as parameter size increases, the WTWF group exhibits sustained growth.
This indicates that monolithic models encounter a saturation point; in contrast, our modular divide-and-conquer strategy mitigates diminishing returns, enabling more efficient capacity utilization at larger scales.

\subsection{Intra-module Analysis}
Our detailed and comprehensive internal comparisons reveal two distinct and significant phenomena.
First, within the NTNF baseline, performance did not scale linearly as one might typically expect; we observe a counter-intuitive drop from 8B to 14B, and Llama3-8B underperformed compared to Qwen3-8B due to inherent language alignment limitations.
Second, a notable “small-beating-large” pattern emerged: the NTWF-4B (zero-shot with framework) achieved a higher All\_in\_One\_Acc than the larger NTNF-8B and NTNF-16B models.
Moreover, within the WTWF group, performance improves rapidly with increasing model scale, confirming that the framework not only raises the performance baseline but also significantly extends the limits of model capability.

\subsection{Generalization on Existing Datasets}
To thoroughly evaluate cross-lingual robustness, we extend our comprehensive experiments to include the InToxiCat (Catalan) and Hate Speech and Offensive Language (English) datasets.
As shown in Table \ref{tab:Performance on other datasets}, the WTWF module maintains a significant performance advantage over NTNF distinct from Chinese contexts.
Specifically, on InToxiCat, WTWF consistently outperforms NTNF across abusive language detection and explicit/implicit analysis tasks.
In addition, the result obtained using our method surpasses the 87.03\% achieved in previous studies of BD-LLM methods \cite{zhang2024efficient}.
Similarly, on the Hate Speech dataset, WTWF achieves a 17.63\% higher recognition accuracy when compared directly to NTNF.
Moreover, our implicit offensive language recognition accuracy exceeds 71.9\%, the best result reported in previous research \cite{wiegand2022identifying}.
These results confirm that our approach possesses strong and reliable generalization capabilities across different languages and cultural domains.

\begin{table}[t]
\centering
\small
\scalebox{0.95}{
\begin{tabular}{@{}lccc@{}}
\toprule
\textbf{Dataset} & \textbf{Strategy} & \textbf{Category} & \textbf{Accuracy} \\
\midrule
\multirow{3}{*}{hate-speech} & NTNF & Binary classification & 0.74 \\
                             & BD-LLM & Binary classification & 0.87 \\
                             & WTWF & Binary classification & 0.92 \\
\midrule
\multirow{7}{*}{IoToxiCat} & \multirow{3}{*}{NTNF} & Abusive & 0.68 \\
                           &                       & Explicit & 0.78 \\
                           &                       & Implicit & 0.68 \\
                           & \multirow{1}{*}{CCBLR} & Implicit & 0.72 \\
                           & \multirow{3}{*}{WTWF} & Abusive & 0.79 \\
                           &                       & Explicit & 0.98 \\
                           &                       & Implicit & 0.81 \\
\bottomrule
\end{tabular}
}
\caption{Performance comparison with previous methods. We benchmark our strategies against BD-LLM and CCBLR on the hate-speech and IoToxiCat datasets, respectively. WTWF consistently achieves superior accuracy across all categories.}
\label{tab:Performance on other datasets}
\vspace{-0.6cm}
\end{table}

\subsection{Case Study}
\begin{figure*}[t] 
    \centering
    \includegraphics[width=1\textwidth, trim=0 5 0 10]{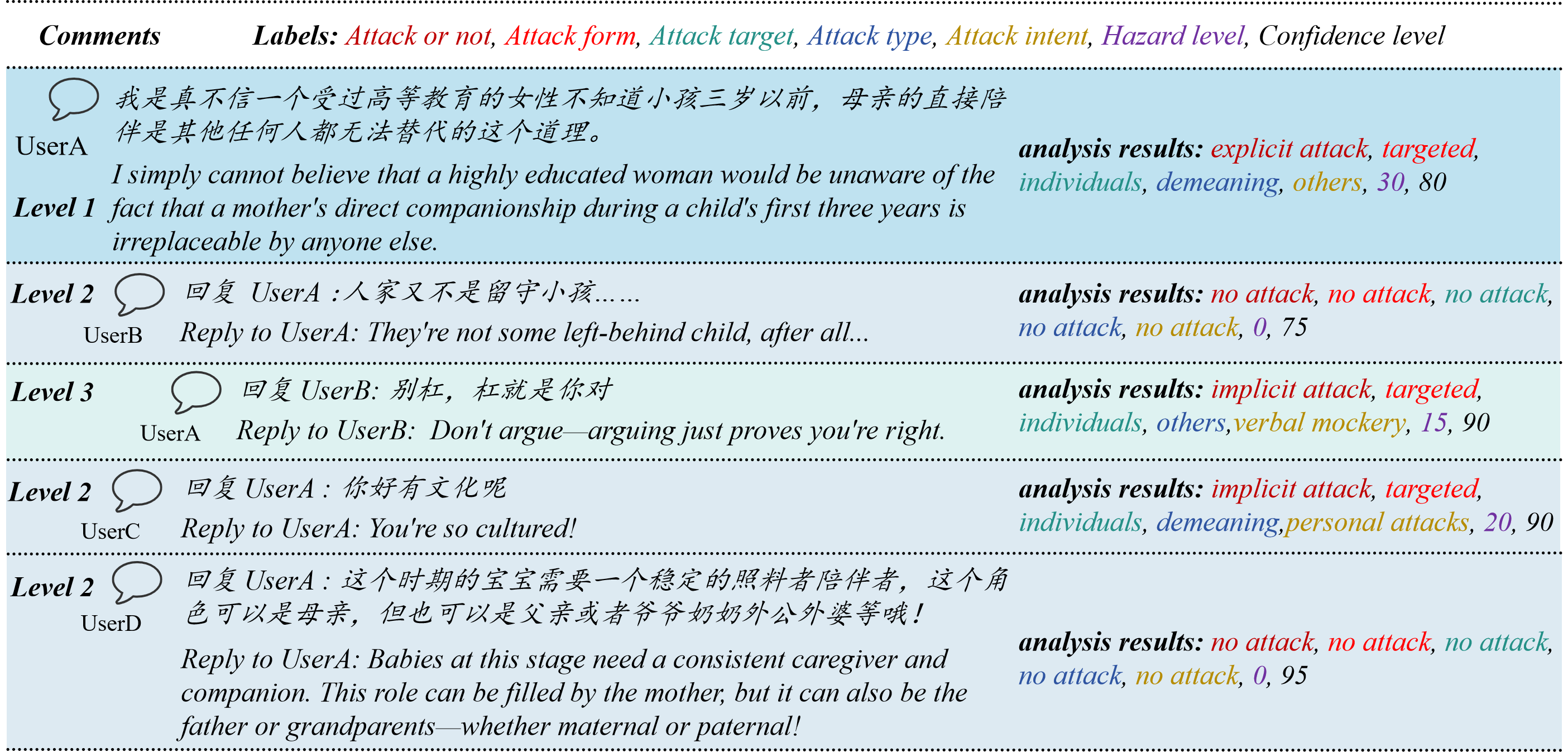}
    \caption{A case study illustrating multi-dimensional analysis on a comment thread. The text on the right displays the model's output for each user comment, covering seven metrics from "Attack or not" to "Confidence level." Note that the color of the text corresponds to the specific label categories defined in the top legend.}
    \label{fig:case study}
    \vspace{-0.6cm}
\end{figure*}

To provide a qualitative assessment of the model's performance, we present a representative case study in Figure \ref{fig:case study}. 
This example demonstrates the framework's capability to interpret complex conversational contexts and accurately distinguish between explicit attack, implicit attack, and no attack.

\section{Related Work}

Social media attack detection has attracted sustained and extensive research attention in recent years, driven by its societal impact and the rapid growth of user-generated online content \cite{hartvigsen2022toxigen,mei2024improving,tian2024learning,tian2025detoxification,usman2025multilingual}.
The detection paradigm has shifted from statistical feature engineering to LLM-based approaches.
Early methods relied on SVMs and manual features \cite{davidson2017automated,biju2017friedman}.
Afterwards, pre-trained language models demonstrate their effectiveness in text processing \cite{devlin2019bert,clark2020electra,qin-etal-2021-relation,chen-etal-2021-relation}, evolving to BERT-based architectures for context capture \cite{mozafari2019bert}. 
Recently, LLMs achieve outstanding performance on data processing \cite{yang2024kg,csahinucc2024systematic,su2025fusing}.
For example, ToxiGen enhances few-shot robustness of LLMs via adversarial generation \cite{hartvigsen2022toxigen}. However, applying general-purpose LLMs in zero-shot settings remains challenging, often yielding high false-positive rates due to the lack of domain-specific instruction tuning \cite{usman2025multilingual}.
Beyond overt toxicity, the field is increasingly focused on the nuanced challenge of implicit attack.
Unlike explicit abuse, implicit forms such as sarcasm lack specific lexical triggers, thereby confusing current models \cite{wiegand2021implicitly,elsherief2021latent}. This complexity drove the creation of granular benchmarks like Latent Hatred \cite{elsherief2021latent} and explainable frameworks like HateXplain, which advocate for generating rationales alongside classification to decode the model's "black box" nature \cite{mathew2021hatexplain}.
Contextual modeling is essential for accuracy, yet existing structure-aware methods often oversimplify conversation dynamics.
While incorporating parent comments limits errors \cite{pavlopoulos2020toxicity} and GNNs effectively model abuse propagation \cite{mishra2019abusive,kumar2022hate}, these approaches typically compress structures into static embeddings. They consequently miss the opportunity to use LLMs for explicit analysis of the tree structure to trace the logical flow of attack target-by-target.

We propose a generative analysis framework to bridge the gap between structural modeling and granular reasoning.
Addressing the conflation of implicit and explicit intent, our approach moves beyond single-label prediction. By leveraging the hierarchical dialogue structure, we enable LLMs to perform fine-grained reasoning, identifying not just the presence of an attack but its dynamic manifestation within the conversation thread.

\section{Conclusion}

In this paper, we propose the Hierarchical Attack Comment Detection (HACD) dataset and a divide-and-conquer framework for fine-grained verbal attack detection. By decomposing the task into hierarchical sub-problems, from explicit detection to implicit intent and target analysis, our approach systematically captures nuanced attacks in online discourse. Leveraging spatiotemporal conversational context, specialized lightweight models collaboratively reason over multi-turn dialogs. Experiments on Chinese and multilingual datasets demonstrate consistent improvements in identifying subtle, context-dependent attacks, with smaller models outperforming larger monolithic baselines in both accuracy and efficiency across diverse conversational settings. Furthermore, our method exhibits strong cross-lingual generalization capabilities, effectively maintaining performance advantages across distinct cultural and linguistic domains. Future work will explore multimodal signals and user behavior to support deployment scenarios and further enhance robustness.

\bibliography{custom}

\end{document}